\documentclass[10pt,twocolumn,letterpaper]{article}

\usepackage{spconf}
\usepackage{epsfig}
\usepackage{graphicx}
\usepackage{amsmath}
\usepackage{amssymb}
\usepackage{algorithm}
\usepackage{booktabs}
\usepackage{algpseudocode}
\usepackage{caption}
\usepackage{xspace}
\usepackage{graphics}
\usepackage{subcaption}
\usepackage{amsfonts}
\usepackage{booktabs}
\usepackage{multirow}
\usepackage{enumitem}
\usepackage{url}
\usepackage{gensymb}
\usepackage{tabularx}

\usepackage[breaklinks=true,bookmarks=false]{hyperref}

\def\cvprPaperID{****} 

\begin{document}

\title{Head Pose Estimation of Occluded Faces\\ 
using Regularized Regression}

\name{Amit Kumar, Rishabh Bindal, Soumya Indela and Michael C. Rotkowitz} 
\address{Department of Electrical and Computer Engineering\\
University of Maryland, College Park, MD 20742\\
{\tt\small\{akumar14, rbindal, sindela, mcrotk\}@umd.edu}
}
\maketitle

\begin{abstract}
This paper presents regression methods for estimation of head pose from occluded 2-D face images. The process primarily involves reconstructing a face from its occluded image, followed by classification. Typical methods for reconstruction assume that the pixel errors of the occluded regions are independent. However, such an assumption is not true in the case of occlusion, because of its inherent contiguous nature. Hence, we use nuclear norm as a metric that can describe well the structure of the error.  We also use LASSO Regression based $l_{1}$ - regularization to improve reconstruction. Next, we implement Nuclear Norm Regularized Regression (NR), and also our proposed method, for reconstruction and subsequent classification. Finally, we compare the performance of the methods in terms of accuracy of head pose estimation of occluded faces. 
\end{abstract}
\section{Introduction}
Estimating the head pose from a 2-D image is important for face recognition and verification, face detection and analysis. Applications such as video surveillance, intelligent environments and human interaction modeling require head pose estimation from low-resolution 2-D face images. The task of pose estimation is particularly more challenging when dealing with occluded images. \\
\indent Most of the available methods in the literature estimate pose by either fitting a 3-D model or by capturing the appearance characteristics using complex neural networks. Keeping in mind the geometrical nature of the problem, we attempt to solve the problem using regression analysis for dictionary based classification. \\
\indent Among appearance-based techniques are methods such as the approach proposed by Meynet et al.\cite{7080569} where a tree of classifiers is trained by hierarchically sub-sampling the pose space, and the technique of Li and Zhang\cite{Li:2004:FLS:1018034.1018354} who apply a detection pyramid which contains classifiers with increasingly finer resolution.  Further appearance-based approaches are the systems developed by Stiefelhagen\cite{DBLP:conf/crv/VoitNS05} and Rae and Ritter\cite{Rae:1998:RHH:2325760.2326229} which are based on neural networks.\\
\indent Model-based approaches use a geometric model of the face for pose estimation. The methods proposed by Stiefelhagen et al.\cite{565083} and Gee and Cipolla\cite{Gee94determiningthe} extract a set of facial features such as eyes, mouth and nose, and map the features onto a 3-D model using perspective projection. The disadvantage with model-based methods is that they are computationally expensive and most of them need manual initialization.\\
\indent In this paper, we propose a method based on \cite{Qian} for pose estimation of occluded images based on regression analysis on a pose dictionary. We follow a two-step process: first, the reconstruction of a face from its occluded image; second, classification into pose categories. We use Nuclear Norm (to capture the structure of occlusion) and $l_{1}$ - norm regularization for better reconstruction of face images with pose variations.

\section{Reconstruction Methods}
In this section, we introduce the Nuclear Norm Regularized Regression (NR) for reconstruction from occluded images, and our proposed modification. These methods code a sample image as a linear combination of the training images. We tailor these reconstruction methods for pose estimation.\\
\indent Suppose that we are given a dataset of \textit{l} matrices $\textbf{A}_{\mathrm{1}},\ldots,\textbf{A}_{l}\in \mathbb{R}^{m\mathrm{x}n}$ and a test matrix $\textbf{Y}\in \mathbb{R}^{m\mathrm{x}n}$. We represent \textbf{Y} linearly by taking the following form:
\begin{equation}
\label{main_eq}
\mathrm{\textbf{Y}} = \mathrm{F(\textbf{x})} + \textbf{E}
\end{equation}
where F(\textbf{x}) = $x_{1}\textbf{A}_{1} + \ldots + x_{l}\textbf{A}_{l}$, \textbf{x} = $(x_{\mathrm{1}}, \ldots, x_{l})^{T} \in \mathbb{R}^{l}$ is the representation coefficient vector, and \textbf{E} is the representation error matrix. The objective, then, is to find a representation coefficient vector \textbf{x} by solving an optimization problem. 

\begin{figure}[b]
\centering
\begin{subfigure}[b]{.25\textwidth} 
\centering 
  \includegraphics[width=\textwidth]{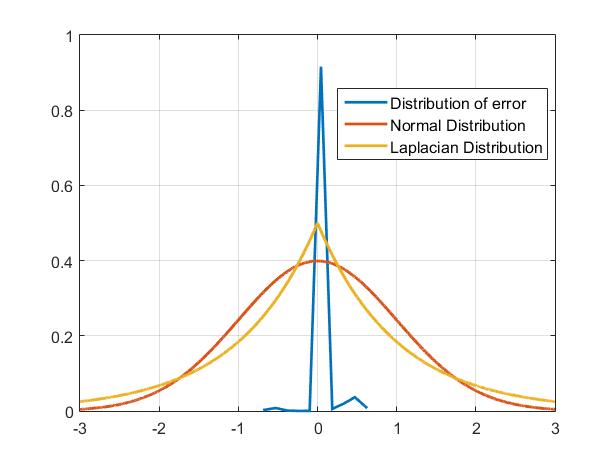}
  \caption{}
  \label{fig:LFPW}
\end{subfigure}%
\begin{subfigure}[b]{.25\textwidth}
\centering 
  \includegraphics[width=\textwidth]{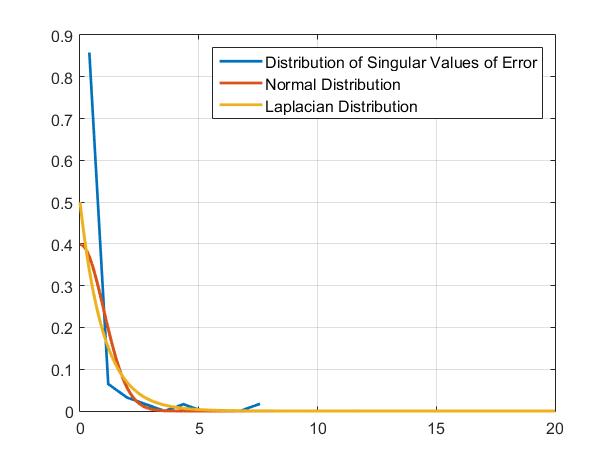}
  \caption{}
  \label{fig:Helen}
\end{subfigure}
\caption{(a) Distribution of Error Image (b) Distribution of Singular Values of Error Image}
\label{fig:plot1}
\end{figure}
\subsection{Nuclear Norm Regularized Regression}

Consider the problem given in (\ref{main_eq}). Regression-based reconstruction methods for occluded images assume that the pixel errors are independent. However, this assumption is not valid in the case of contiguous occlusion, as the errors are spatially correlated. Also, in this case, the error image is neither sparse, nor low rank (for the example shown in Figure~\ref{fig:plot1}, the error image of size 165 x 120 was found to have full column rank). To this end, Qian et al. \cite{Qian} proposed a Nuclear Norm Regularized Regression (NR)-based algorithm to find an optimal representation coefficient vector \textbf{x}.\\ 
\indent In general, the $l_{1}$ - norm best describes the error image when it follows a Laplacian distribution, while the $l_{2}$ - norm is useful for the case of Gaussian distribution. As shown in Figure~\ref{fig:LFPW}, the distribution of a structurally correlated error image does not follow either of the distributions. Hence, $l_{1}$ and $l_{2}$ norms cannot characterize this kind of occlusion effectively. From Figure~\ref{fig:Helen}, it can be seen that the singular values of error image fit the Laplacian distribution. Since the singular values are non-negative, and the nuclear norm is the sum of singular values of a matrix, so it can be considered as $l_{1}$ - norm of the singular value vector. Therefore, the nuclear norm has been utilized in this algorithm to capture the structure of error. Additionally, an $l_{2}$ - norm based regularization term for \textbf{x} has been added to avoid overfitting.\\
\indent Based on this, the method is aimed at finding the optimum representation coefficient vector \textbf{x} by solving the following NR optimization problem:
\begin{equation}
\label{NR_2norm}
\begin{aligned}
&\underset{\textbf{x}}{\text{minimize}}
& \mathrm{\parallel{F(\textbf{x})-\textbf{Y}}\parallel^{2}_{F}} + \lambda \mathrm{\parallel{F(\textbf{x})-\textbf{Y}}\parallel_{*}} + \frac{\eta}{2}\parallel\textbf{x}\parallel^{2}_{2} 
\end{aligned}
\end{equation}
where $\mathrm{F}(\textbf{x}) = \sum\limits_{i=1}^{l} x_{\mathrm{i}} \textbf{A}_{\mathrm{i}}$ \\ $\lambda$ and $\eta$ are regularization parameters and $\textbf{x}\in\mathbb{R}^{l}$ is the representation coefficient vector.\\ \\
\textbf{Solving the NR Problem}:\\
The problem described in (\ref{NR_2norm}) can be reformulated as:
\begin{equation}
\label{NR}
\begin{aligned}
& \underset{\textbf{x},\textbf{E}}{\text{minimize}}
& & \mathrm{\parallel{\textbf{E}}\parallel^{2}_{F}} + \lambda \mathrm{\parallel{\textbf{E}}\parallel_{*}} + \frac{\eta}{2}\parallel\textbf{x}\parallel^{2}_{2} \\
& \text{subject to}
& & \mathrm{F}(\textbf{x})-\textbf{Y} = \textbf{E} \\
\end{aligned}
\end{equation}
 We solve the above problem using the Alternating Direction Method of Multipliers (ADMM).\\
\indent The augmented Lagrangian for (\ref{NR}) is given by :
\begin{equation}
\begin{split}
\begin{aligned}
\label{ALNR}
L_{\mu}(\textbf{x},\textbf{E},\textbf{Z}) = \mathrm{\parallel{\textbf{E}}\parallel^{2}_{F}} + \lambda\mathrm{\parallel{\textbf{E}}\parallel_{*}} + \frac{\eta}{2}\mathrm{\textbf{x}}^{T}\mathrm{\textbf{x}} + \\
\mathrm{Tr}(\textbf{Z}^{T}(\mathrm{F}(\textbf{x})-\textbf{E}-\textbf{Y}) + \\
\frac{\mu}{2}\parallel \mathrm{F}(\textbf{x})-\textbf{E}-\textbf{Y}\parallel^{2}_{\mathrm{F}} 
\end{aligned}
\end{split}
\end{equation}
where $\mu>0$ is the penalty parameter, \textbf{Z} is the Lagrange multiplier and Tr($\cdot$) is the trace operator. \\
\indent ADMM consists of the following iterations:
\begin{equation}
\label{xkplusone}
\mathrm{\textbf{x}}^{k+1} = \mathrm{arg\underset{\textbf{x}}min}\hskip4pt L_{\mu}(\mathrm{\textbf{x}})
\end{equation}
\begin{equation}
\label{Ekplusone}
\mathrm{\textbf{E}}^{k+1} = \mathrm{arg\underset{\textbf{E}}min}\hskip4pt L_{\mu}(\mathrm{\textbf{E}})
\end{equation}
\begin{equation}
\label{Zkplusone}
\mathrm{\textbf{Z}}^{k+1} = \mathrm{\textbf{Z}}^k + \mu(\mathrm{F(\textbf{x}}^{k+1})-\mathrm{\textbf{E}}^{k+1} -\mathrm{\textbf{Y}})
\end{equation} 
\textbf{\textit{Updating x: }}The function $L_{\mu}(\textbf{x})$ in (\ref{xkplusone}) is given by:
\begin{equation}
\label{l_x}
L_{\mu}(\textbf{x})=\frac{\eta}{2}\mathrm{\textbf{x}}^{T}\mathrm{\textbf{x}}+\mathrm{Tr}(\textbf{Z}^{T}(\mathrm{F}(\textbf{x}))+\frac{\mu}{2}\parallel \mathrm{F}(\textbf{x})-\textbf{E}-\textbf{Y}\parallel^{2}_{\mathrm{F}}
\end{equation}
\indent We define $\textbf{H}=[\mathrm{Vec}(\textbf{A}_{1}),\ldots,\mathrm{Vec}(\textbf{A}_{n})]$, $\textbf{g} = \mathrm{Vec}(\mathrm{\textbf{E}}^{k}+\mathrm{\textbf{Y}}-\frac{1}{\mu}\textbf{Z})$ where Vec($\cdot$) converts a matrix into a vector. The above problem can then be simplified as a ridge regression model, the solution to which is shown below:  
\begin{equation}
\label{x_2}
\begin{split}
\mathrm{\textbf{x}}^{k+1}=(\mathrm{\textbf{H}}^{T}\mathrm{\textbf{H}}+\frac{\eta}{\mu}\textbf{I})^{-1}\mathrm{\textbf{H}}^{T}\textbf{g}
\end{split}
\end{equation} 
\textbf{\textit{Updating \textbf{E: }}}The function $L_{\mu}(\textbf{E})$ in (\ref{Ekplusone}) can then be rewritten as:
\begin{equation}
\label{l_e}
L_{\mu}(\textbf{E}) = \parallel\mathrm{\textbf{E}}\parallel^{2}_{\mathrm{F}} +\lambda\parallel \mathrm{\textbf{E}}\parallel _{*}-\mathrm{Tr}(\mathrm{\textbf{Z}}^{T}\mathrm{\textbf{E}})+\frac{\mu}{2}\parallel \mathrm{F}(\textbf{x})-\textbf{E}-\textbf{Y}\parallel^{2}_{\mathrm{F}}
\end{equation}
\indent The solution to the above problem is given by: 
\begin{equation}
\label{e_2}
\mathrm{\textbf{E}}^{k+1} = \mathrm{\textbf{U}}T_{\frac{\lambda}{\mu+2}}[\textbf{S}]\textbf{V}
\end{equation}
where (\textbf{U},\textbf{S},$\textbf{V}^{T}) = \mathrm{svd}(\frac{\mu}{\mu+2}(\mathrm{F}(\textbf{x})-\textbf{Y}+\frac{1}{\mu}\textbf{Z})$.\\
\par The singular value shrinkage operator $T_{\frac{\lambda}{\mu+2}}[\textbf{S}]$ is defined as $T_{\frac{\lambda}{\mu+2}}[\textbf{S}]=\mathrm{diag}(\{\mathrm{max}(0,s_{jj}-\frac{\lambda}{\mu+2})\}_{1\leq j \leq r })$, where \textit{r} is the rank of \textbf{S}.\\
\indent Given a set of matrices $\textbf{A}_{1}$,\ldots,$\textbf{A}_{l}$ and a matrix $\textbf{Y} \in$ $\mathbb{R}^{m\mathrm{x}n}$, the model parameters $\lambda$ \& $\eta$, and the termination condition parameter $\epsilon$; $\textbf{E}^{0}$, $\textbf{Z}^{0}$ \& $\mu$ are initialized; and \textbf{x}, \textbf{E} \& \textbf{Z} are updated until the following stopping criteria : \\
$\parallel \mathrm{F}(\textbf{x}^{k+1})-\textbf{E}_{k+1}-\textbf{Y}\parallel_{\mathrm{F}}\leq\epsilon$ or  \\
$\mathrm{max}(\parallel \textbf{x}^{k+1}-\textbf{x}^{k}\parallel_{\mathrm{F}},\parallel \textbf{E}^{k+1}-\textbf{E}^{k}\parallel_{\mathrm{F}})\leq\epsilon$ is met. \\
\indent We direct the readers to \cite{Qian} for more insight into the algorithm and its convergence.
\subsection{Proposed Modification - NR with $l_{1}$ Regularization}
The method discussed in the previous section uses $l_{2}$ – regularization (ridge regression) for penalizing the representation coefficient vector \textbf{x}. Ridge regression penalizes the components of \textbf{x}, without ever driving them to $0$. However, in our case, it is desired that \textbf{x} be of the form $[0,...,0, \alpha_{i1},..., \alpha_{it}, 0,..., 0]^{T}$, with non-zero coefficients corresponding to the dictionary columns from the $i^{th}$ class, to which the test image belongs. \\
To tackle this problem, we turn to the LASSO regression algorithm as described in \cite{Tibshirani94regressionshrinkage}. LASSO uses $l_{1}$ – regularization to minimize least squares. An increase in penalty associated with the $l_{1}$ – norm forces some coefficients to become $0$, which can be advantageous for our problem. Based on this, we propose using $l_{1}$ – regularization for \textbf{x} instead of $l_{2}$ – regularization in (\ref{NR}). Using this, we get the modified optimization problem as:
\begin{equation}
\label{l1NR}
\begin{aligned}
& \underset{\textbf{x},\textbf{E}}{\text{minimize}}
& & \mathrm{\parallel{\textbf{E}}\parallel^{2}_{F}} + \lambda \mathrm{\parallel{\textbf{E}}\parallel_{*}} + \frac{\eta}{2}\parallel\textbf{x}\parallel_{1} \\
& \text{subject to}
& & \mathrm{F}(\textbf{x})-\textbf{Y} = \textbf{E} \\
\end{aligned}
\end{equation}
where $\mathrm{F}(\textbf{x})$, \textbf{x}, \textbf{Y} and $\textbf{A}_{i}$, $i=1,\ldots,l$ are as defined in the previous section. \\
The augmented Lagrangian for (\ref{l1NR}) is given by :
\begin{equation}
\label{AL1NR}
\begin{split}
\begin{aligned}
L_{\mu}(\textbf{x},\textbf{E},\textbf{Z}) = \mathrm{\parallel{\textbf{E}}\parallel^{2}_{\mathrm{F}}} + \lambda\mathrm{\parallel{\textbf{E}}\parallel_{*}} + \frac{\eta}{2}\parallel\textbf{x}\parallel_{1} +\\
\mathrm{Tr}(\textbf{Z}^{T}(\mathrm{F}(\textbf{x})-\textbf{E}-\textbf{Y}) + \\ 
\frac{\mu}{2}\parallel \mathrm{F}(\textbf{x})-\textbf{E}-\textbf{Y}\parallel^{2}_{\mathrm{F}}
\end{aligned}
\end{split}
\end{equation}
where $\mu>0$ is the penalty parameter, \textbf{Z} is the Lagrange multiplier and Tr(.) is the trace operator.\\ \\
\textbf{Solving the Modified NR Problem}:\\
\indent We again use the Alternating Direction Method of Multipliers (ADMM) for solving the given problem. The update equations for ADMM are same as described in (\ref{xkplusone}), (\ref{Ekplusone}) and (\ref{Zkplusone}).\\
\textbf{\textit{Updating x: }}The function $L_{\mu}(\textbf{x})$ in (\ref{xkplusone}) is now:
\begin{equation}
\label{l1_x}
L_{\mu}(\textbf{x}) = \frac{\eta}{2}\parallel\textbf{x}\parallel_{1} + \mathrm{Tr}(\textbf{Z}^{T}(\mathrm{F}(\textbf{x})) + \frac{\mu}{2}\parallel \mathrm{F}(\textbf{x})-\textbf{E}-\textbf{Y}\parallel^{2}_{\mathrm{F}}
\end{equation}
\indent The corresponding update equation is given by:
\begin{equation}
\label{x1_2}
\begin{split}
\mathrm{\textbf{x}}^{k+1} = \mathrm{arg\underset{\textbf{x}}min}\hskip4pt (\frac{\mu}{2}\parallel\mathrm{\textbf{H}\textbf{x}-\textbf{g}}\parallel^{2}_{2} + \frac{\eta}{2}\parallel\textbf{x}\parallel_{1})
\end{split}
\end{equation} 
\indent where \textbf{H} and \textbf{g} are as defined in the previous section. \\
\indent The solution to the problem described above can be obtained using proximal operator as: 
\begin{equation}
\label{x1_2}
\begin{split}
\mathrm{\textbf{x}}^{k+1} = \mathrm{prox}_{\frac{\eta}{2\mu}} (\mathrm{\textbf{H}}^{T}\mathrm{\textbf{H}})^{-1}\mathrm{\textbf{H}}^{T}\textbf{g}) 
\end{split}
\end{equation} 
\indent Denoting the quantity $(\mathrm{\textbf{H}}^{T}\mathrm{\textbf{H}})^{-1}\mathrm{\textbf{H}}^{T}\textbf{g}$  by  $\boldsymbol\alpha\in\mathbb{R}^{l}$, the update equation for the $i^{th}$ component of \textbf{x} is given by:
\begin{equation}
\label{x1_2}
\begin{split}
\mathrm{\textbf{x}}^{k+1}_{i} = \begin{cases} \boldsymbol\alpha_{\mathrm{i}} - \frac{\eta}{2\mu}, & \mbox{if } \boldsymbol\alpha_{\mathrm{i}} > \frac{\eta}{2\mu} \\ 0, & \mbox{if } |\boldsymbol\alpha_{\mathrm{i}} | \leq \frac{\eta}{2\mu} \\ \boldsymbol\alpha_{\mathrm{i}} + \frac{\eta}{2\mu}, & \mbox{if } \boldsymbol\alpha_{\mathrm{i}} < -\frac{\eta}{2\mu} \end{cases}
\end{split}
\end{equation}
\textbf{\textit{Updating \textbf{E} and \textbf{Z: }}}The function $L_{\mu}(\textbf{E})$ remains the same as in (\ref{l_e}). Thus, the corresponding update equations for \textbf{E} and \textbf{Z} remain the same as in the previous section ((\ref{e_2}), (\ref{Zkplusone})). We use the same stopping criteria as in the previous section.

\section{Experimental Details and Results}
\begin{figure}[t]
\begin{subfigure}[t]{0.5\textwidth}
  \includegraphics[width=0.5\textwidth]{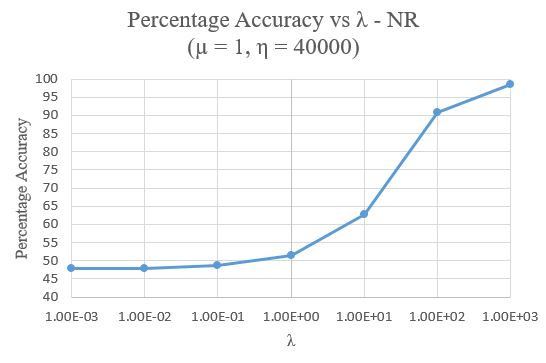}\includegraphics[width=0.5\textwidth]{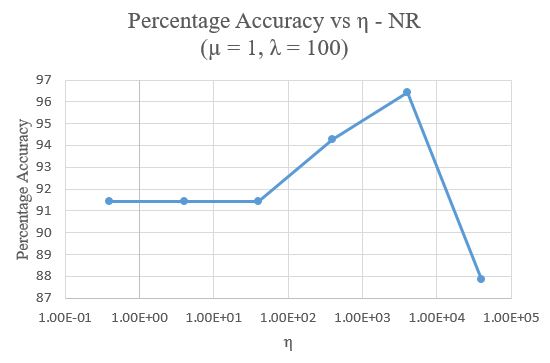}
\end{subfigure}
\caption{Accuracy \textit{vs} parameters $\lambda$ and $\eta$ for NR}
\label{fig:plot5}
\end{figure}
\begin{figure}[t]
\begin{subfigure}[t]{0.5\textwidth}
  \includegraphics[width=0.5\textwidth]{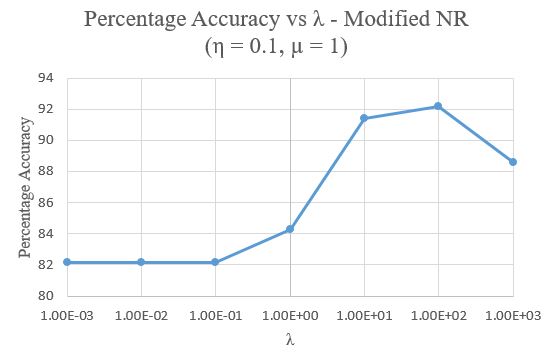}\includegraphics[width=0.5\textwidth]{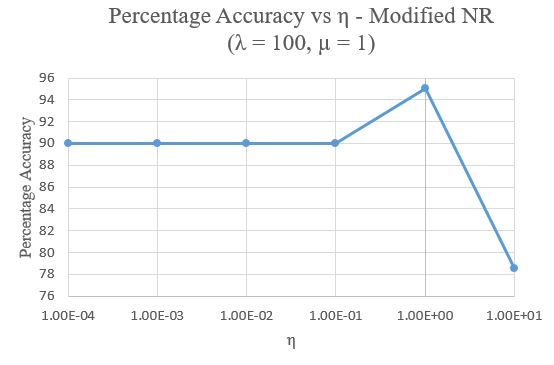}
\end{subfigure}
\caption{Accuracy \textit{vs} parameters $\lambda$ and $\eta$ for Modified-NR}
\label{fig:plot6}
\end{figure}

\begin{figure*}[t]

\hskip30pt\begin{subfigure}{0.1\textwidth}
\centering
\includegraphics[width=1.5cm,height=1.5cm]{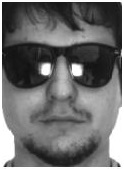}\caption{}
\end{subfigure}
\begin{subfigure}{0.9\textwidth}
\centering
\includegraphics[width=1.5cm]{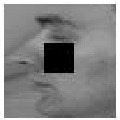}\hskip4pt\includegraphics[width=1.5cm]{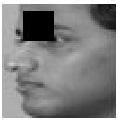}\hskip4pt\includegraphics[width=1.5cm]{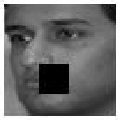}\hskip4pt\includegraphics[width=1.5cm]{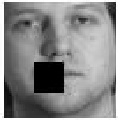}\hskip4pt\includegraphics[width=1.5cm]{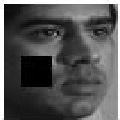}\hskip4pt\includegraphics[width=1.5cm]{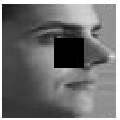}\hskip4pt\includegraphics[width=1.5cm]{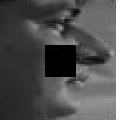}\caption{}

\end{subfigure}

\hskip30pt\begin{subfigure}{0.1\textwidth}
\centering
\includegraphics[width=1.5cm,height=1.5cm]{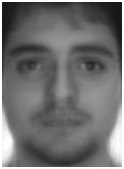}\caption{}
\end{subfigure}
\begin{subfigure}{0.9\textwidth}
\begin{center}
\includegraphics[width=1.5cm]{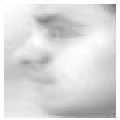}\hskip4pt\includegraphics[width=1.5cm]{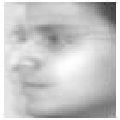}\hskip4pt\includegraphics[width=1.5cm]{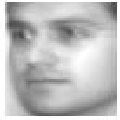}\hskip4pt\includegraphics[width=1.5cm]{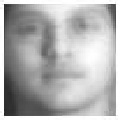}\hskip4pt\includegraphics[width=1.5cm]{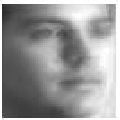}\hskip4pt\includegraphics[width=1.5cm]{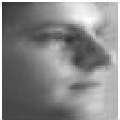}\hskip4pt\includegraphics[width=1.5cm]{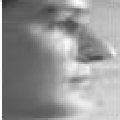}\caption{}
\end{center}
\end{subfigure}

\hskip30pt\begin{subfigure}{0.1\textwidth}
\centering
\includegraphics[width=1.5cm,height=1.5cm]{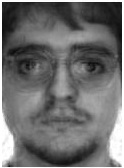}\caption{}
\end{subfigure}
\begin{subfigure}{0.9\textwidth}
\begin{center}
\includegraphics[width=1.5cm]{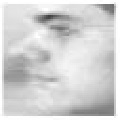}\hskip4pt\includegraphics[width=1.5cm]{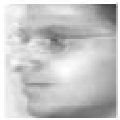}\hskip4pt\includegraphics[width=1.5cm]{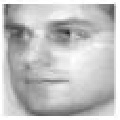}\hskip4pt\includegraphics[width=1.5cm]{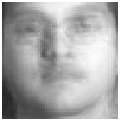}\hskip4pt\includegraphics[width=1.5cm]{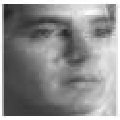}\hskip4pt\includegraphics[width=1.5cm]{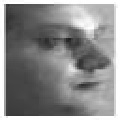}\hskip4pt\includegraphics[width=1.5cm]{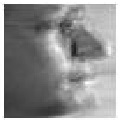}\caption{}
\end{center}
\end{subfigure}
\caption{(a) Sample image from \cite{AR} and (b) test images in angles Left-$90\degree$, Left-$60\degree$, Left-$30\degree$, Frontal, Right-$30\degree$, Right-$60\degree$, Right-$90\degree$ with block occlusion, (c) and (d): Corresponding reconstructions by NR, (e) and (f): Corresponding reconstructions by Modified-NR, capturing more facial details than (c) and (d)}
\label{fig:images}
\end{figure*}
\begin{figure}[H]
\begin{subfigure}[htp!]{0.5\textwidth}
\begin{center}
  \includegraphics[width=\textwidth,height=5cm]{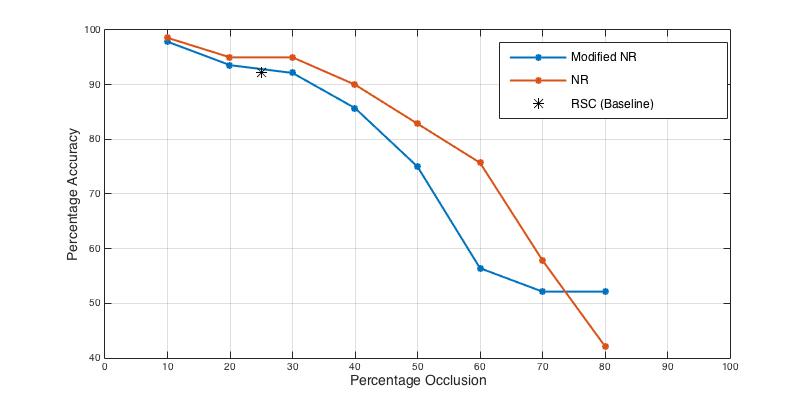}
  \label{LFPW}
  \end{center}
\end{subfigure}
\caption{Accuracy \textit{vs} Occlusion for NR and Modified-NR compared with the baseline method RSC \cite{5995393}}
\label{fig:plot7}
\end{figure}

\indent We use the algorithms to classify the images based on a set of pose angles. For this purpose, we use the MULTIPIE dataset \cite{Gross:2010:MUL:1746745.1747071}. It contains 337 subjects, captured under 15 view points and 19 illumination conditions in four recording sessions for a total of more than 750,000 images.\\
\indent We construct our set of training images $\textbf{A}_{i}$'s (vectorized to form \textbf{D}) by using a subset of the dataset, consisting of 50 images for each of the pose angles $0^{\mathrm{o}}$, $\pm 30^{\mathrm{o}}$, $\pm 60^{\mathrm{o}}$ and $\pm 90^{\mathrm{o}}$ for the training set. The training dictionary, thus, comprises of $50$ images each from $7$ pose classes. We use 20 images of each of the pose angles for the test set. Block occlusion is added to the test images by randomly blackening the pixels in a square area of specified size. Both the training and test images are cropped to a size of $64 \times 64$. Using the algorithms described in Section 2, the corresponding optimum representation coefficient vectors are obtained for each of the test images. Using the coefficients corresponding to each particular pose class, $7$ corresponding reconstructions are obtained for each test image. Reconstruction residuals from each class are computed using the Frobenius norm, and the pose class is assigned based on the minimum residual error. The percentage accuracy for an algorithm is calculated as the percentage of test images correctly classified.\\ \\
\textbf{Reconstruction using NR and Modified-NR}: As seen in section 2, the ADMM iterations for NR and Modified-NR depend on the parameters $\lambda$, $\eta$ and $\mu$. We vary these parameters to obtain the optimal parameter set for which the pose estimation accuracy is maximum. For NR, $\lambda$ is varied from $0.001$ to $1000$ in powers of $10$, for fixed $\eta$ = $40,000$ and $\mu$ = $1$. Then, $\eta$ is varied from $0.4$ to $40,000$ in powers of $10$ for fixed $\lambda$ = $100$ and $\mu$ = 1. The corresponding plots are shown in Figure~\ref{fig:plot5}. Next, $\mu$ is varied to be $0.1$, $1$ and $10$ for fixed $\eta = 4000$ and $\lambda = 100$. Although from Figure~\ref{fig:plot5}, it can be seen that $\lambda = 1000$ gives the highest accuracy, a value of $\lambda = 100$ is chosen to speed up the process. An accuracy of $97.14\%$ is obtained for all the values of $\mu$ under this setting. Thus, the optimal parameters for NR are found to be $\lambda = 100$, $\eta = 40,000$ and $\mu = 1$.\\
\indent Similarly, for Modified-NR, $\lambda$ is varied from $0.001$ to $1000$ in powers of $10$ for fixed values of $\eta$ = $0.1$ and $\mu$ = $1$. Next, $\eta$ is varied from $0.0001$ to $10$ in powers of $10$ for fixed values of $\lambda$ = $100$ and $\mu$ = $1$. The corresponding plots are shown in Figure~\ref{fig:plot6}. Finally, $\mu$ is varied to be $1$, $10$ and $100$  for fixed $\eta$ = $0.1$ and $\lambda$ = $100$. The values of accuracy obtained are $92.86\%$ for $\mu$ = $1$, $91.43\%$ for $\mu = 10$ and $92.14\%$ for $\mu = 100$. From the values, it can be seen that the accuracy does not vary much for variation in $\mu$. Thus, the optimal parameters for Modified-NR are found to be $\lambda = 100$, $\eta = 1$ and $\mu = 1$.\\ \\
\textbf{Variation in Percentage Occlusion}: After obtaining the set of optimal parameters, the percentage of occlusion in the test images is varied from $10\%$ to $80\%$ in steps of 10 along each image axis. Figure~\ref{fig:images} shows the reconstructed images with correct pose classification from $25\%$ occluded images. Figure~\ref{fig:plot7} shows the percentage accuracies for both NR and Modified-NR by varying the percentage occlusion. As expected, the accuracy decreases with increasing occlusion for both the algorithms. Although the accuracy of the modified NR is less, the reconstructed image captures more details. 
\section{Conclusion and Future Work}
In this paper, we propose a modification of existing methods of reconstruction from occluded images, tailored for addressing our problem. Extensive experiments demonstrate the effectiveness of regression based methods for pose estimation. The advantage of accounting for the structure of error for accurate reconstruction is demonstrated. Finally, the experiments show the effects of parameter selection, percentage occlusion and type of algorithm on the performance of the methods.\\
\indent In our future work, we will consider methods for improved parameter selection, such as the use of solutions obtained in the previous iterations for speeding up future iterations. Additionally, we will consider methods to increase the number of pose angles that can be correctly classified.


{\small
\bibliographystyle{ieee}
\bibliography{report_bib}
}

\end{document}